\def\year{2016}\relax
\pdfoutput=1
\documentclass[letterpaper]{article}
\usepackage{aaai16}
\usepackage{times}
\usepackage{helvet}
\usepackage{courier}
\usepackage{graphicx}
\graphicspath{ {images/} }
\nocopyright
\begin{document}
%
\title{Ashwin: Plug-and-Play System for Machine-Human Image Annotation}

\author{A. Sriraman, M. Kulkarni, R. Kumar, K. Kalra,  P. Radadia, S. Karande\\
TCS Research, Tata Consultancy Services, Pune, India\\
\{anand.sriraman, mandar.kulkarni3, rahul.14, kalra.kanika, purushotam.radadia, shirish.karande\}@tcs.com
}

\maketitle
\begin{abstract}
We present an end-to-end machine-human image annotation system where each component can be attached in a plug-and-play fashion. These components include Feature Extraction, Machine Classifier, Task Sampling and Crowd Consensus.
\end{abstract}

\section{Introduction}

\noindent The rise of online labour marketplaces like Amazon's Mechanical Turk, CrowdFlower, etc. has spurred research to use crowd annotation for training machine learning algorithms. Modern ML algorithms require large amounts of annotated data and hence active learning \cite{mozafari2014scaling} and machine-human collaborative systems \cite{russakovsky2015best} have been investigated.

Different algorithms have their advantages depending on the application. Hence, applying the right algorithm for the problem is crucial. Towards this goal, we seek to build a plug-and-play system for machine-human image annotation where users can upload their own algorithms and choose which ones to use for a specific job.

\section{System Workflow}

\subsection{Workflow}

\begin{figure} [!h]
\centering
\begin{tabular}{c}
\includegraphics[height = 120pt,width = 230pt]{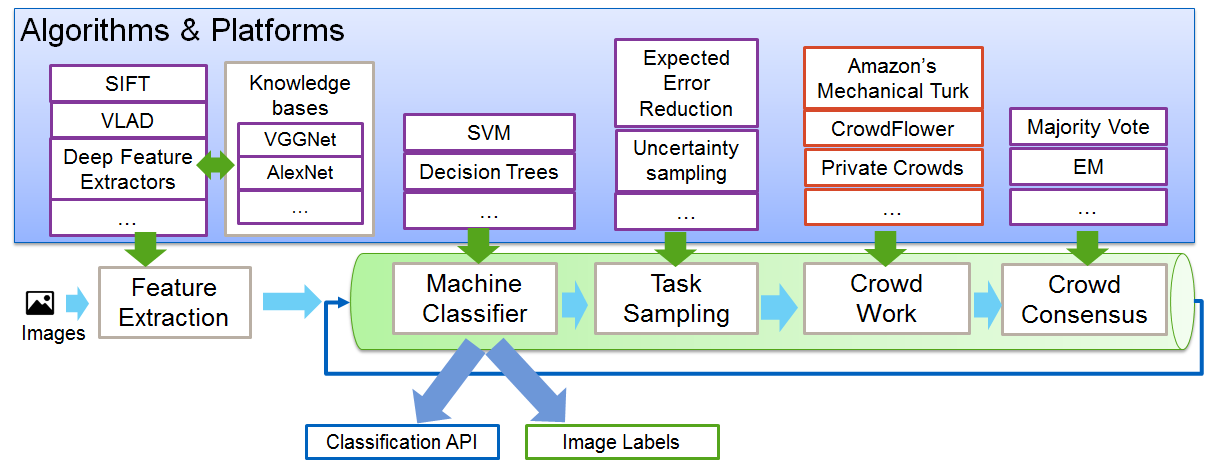}\
\end{tabular}
\caption{\label{fig:pipeline} System Workflow}
\end{figure}

Figure \ref{fig:pipeline} shows the entire workflow of our system and is described below. Throughout this paper, we differentiate between the requester (person who wants the images annotated), the researcher (person who uploads programs for different parts of the pipeline) and the workers (people who annotate the images).

\begin{itemize}
\item An image dataset is fed to \textbf{Feature Extraction} to extract a feature vector for the images.
\item The \textbf{Machine Classifier} is trained on a seed dataset. This is used to classify and provide confidence values for the all the extracted feature vectors.
\item The \textbf{Task Sampling} stage uses the classification confidence values to determine which images to get labelled.
\item \textbf{Crowd Work} is performed on the sampled images and a \textbf{Crowd Consensus} is formed.
\end{itemize}

The consensus labels are fed back to the \textbf{Machine Classifier} for re-training and the loop continues.

At each stage in the above workflow, the algorithm to be used can be changed. For example, Feature Extraction could be a traditional feature extractor like SIFT or it can be a deep network-based extractor \cite{sharif2014cnn}. The Classifier could be SVM, Decision Tree, etc. or a deep learning based approach \cite{krizhevsky2012imagenet}. Task Sampling for Active Learning can use one of the many algorithms proposed in the field, e.g. uncertainty sampling \cite{cohn1996active}. Crowd consensus could be a simple majority vote or an EM-based approach \cite{welinder2010online}. 

\subsection{Job Configurability}

\begin{figure} [!h]
\centering
\begin{tabular}{c}
\includegraphics[height = 90pt,width = 230pt]{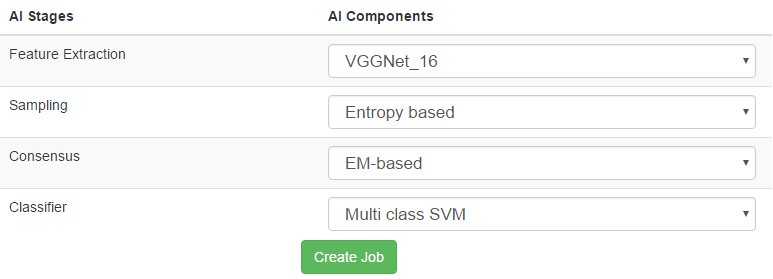}\
\end{tabular}
\caption{\label{fig:mapping} Job Component Configuration UI}
\end{figure}

Requesters upload their Job, consisting of the image dataset to be labelled, type of label required (class/bounding box) and an initial training dataset. They can select which algorithm, from those uploaded by researchers, to use for each machine stage (see Figure \ref{fig:mapping}). They can also choose whether to recruit public (MTurk/CrowdFlower) or private crowds.

\section{Algorithm Configurability}

\subsection{Algorithm Upload}
Researchers upload their algorithm code for any stage of the above pipeline. These programs can be marked public (visible to all) or private (visible to only the uploader). Programs uploaded by researchers are verified by authorized users to make sure they are safe to run.

Researchers can program their algorithm implementations in Python and upload a ZIP file containing all required files, including any model files. They must specify the file containing methods matching the signature specified below:

\begin{itemize}
\item Feature Extraction
\begin{itemize}
\item \textit {getModel()} : model object
\item \textit {getFeatureVector(image, model)} : float list
\end{itemize}
\item Machine Classification
\begin{itemize}
\item \textit {doTrain(images, image\_labels)} : model object
\item \textit {doRun(image,model)} : image\_label
\end{itemize}
\item Task Sampling
\begin{itemize}
\item \textit {getNextSamples(images,image\_labels)} : image list
\end{itemize}
\item Crowd Consensus
\begin{itemize}
\item \textit {getConsensus(images,crowd\_labels)} : consensus\_labels
\end{itemize}
\end{itemize}

\begin{figure} [!h]
\centering
\begin{tabular}{c}
\includegraphics[height = 85pt,width = 230pt]{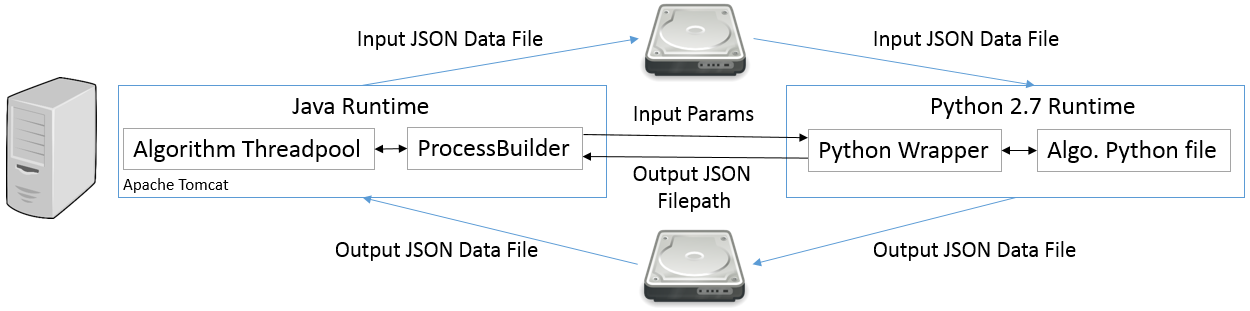}\
\end{tabular}
\caption{\label{fig:architecture} Backend Architecture}
\end{figure}

Figure \ref{fig:architecture} shows the backend architecture. Our server application is developed using Java. Our Python environment has relevant packages installed for Deep Learning as well, e.g. Theano, TensorFlow, etc. The Python programs run as an asynchronous external process since program execution may not be real-time. The communication between the Java and Python processes happens through temporary JSON files.

\section{Crowd Work}

\subsection{Crowd Annotation}

\begin{figure} [!h]
\centering
\begin{tabular}{c}
\includegraphics[height = 100pt,width = 230pt]{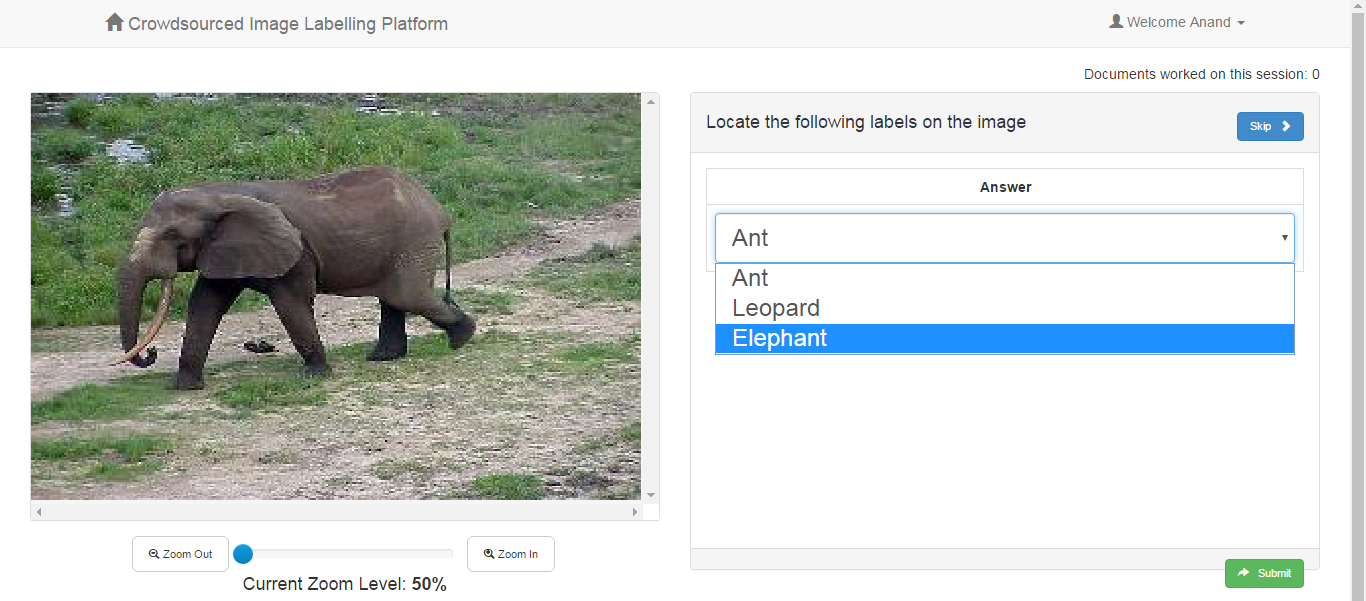}\
\end{tabular}
\caption{\label{fig:crowd-ui} Worker UI for Classification}
\end{figure}

Supported annotation types by the crowd are: (i) Classification, (ii) Bounding Box, (iii) Object Contour and (iv) Image Comparison (Are two images same?). For each annotation type, the crowd worker will be shown the corresponding user interface for performing the task. A screenshot of the user interface for the classification annotation type is shown in Figure \ref{fig:crowd-ui}. Crowds for performing the work are recruited using the Coordination Layer.

\subsection{Coordination Layer}

The coordination layer is responsible for posting HITs on crowd platforms and recruiting workers for batches as generated by the Task Sampling stage. It generates a unique URL for each batch where workers can come to perform the work. If the job is configured for Private Crowds, then this URL is shown to the Requester, who can share it with their trusted crowd. If the job is configured to allow public crowds, then the layer posts a Survey Job on MTurk and CrowdFlower. Crowd workers can then navigate to the given URL and work on the custom UI provided by the system. The coordination layer also takes care of generating the survey code for workers and platform specific timers (CrowdFlower has a 30 minute limit for survey jobs).

\section{On-Demand Web API for Classification}

Every time the Machine Classifier is trained in the above workflow, an on-demand Web API is generated that can be used to run the classifier. If the requester feels that the results are not good enough, they can request for another batch of crowd annotations. This will trigger the Task Sampling stage and the resultant consensus crowd annotations will be used to re-train the classifier. The generation of on-demand classification APIs allows researchers and developers to quickly build applications for new use cases.

\section{Use Case}

We used our pipeline to build a barcode localization system. The entire workflow was configured to train a 1-class SVM for barcodes. The generated API was then used to classify regions on a query image. The barcode was localized to the region with the highest classifier score. 

\begin{figure} [!h]
\centering
\begin{tabular}{c}
\includegraphics[height = 80pt,width = 230pt]{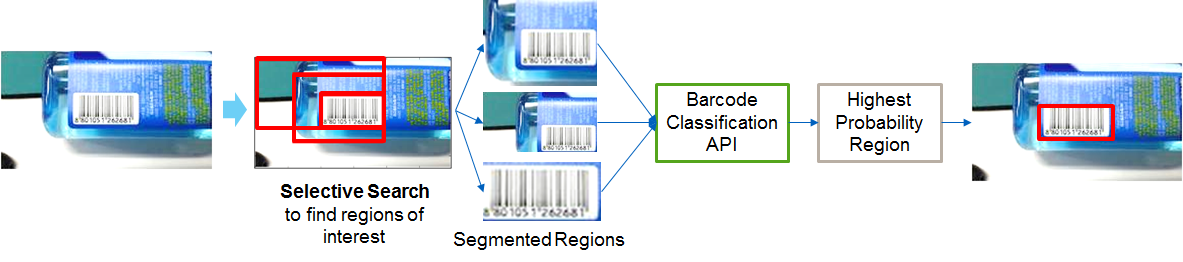}\
\end{tabular}
\caption{\label{fig:barcode} Barcode Localization Workflow}
\end{figure}


\bibliographystyle{aaai}
\bibliography{HCOMP-2016-Ashwin}

\end{document}